\documentclass[conference]{IEEEtran}
\IEEEoverridecommandlockouts
\usepackage{cite}
\usepackage{amsmath,amssymb,amsfonts}
\usepackage{algorithmic}
\usepackage[dvipdfmx]{graphicx}
\usepackage[dvipdfm]{color}
\usepackage{graphicx}
\usepackage{textcomp}
\usepackage{xcolor}
\usepackage{algorithm, algorithmic}
\usepackage{subcaption} 

\def\BibTeX{{\rm B\kern-.05em{\sc i\kern-.025em b}\kern-.08em
    T\kern-.1667em\lower.7ex\hbox{E}\kern-.125emX}}
\begin{document}

\title{Scalable Unsupervised Segmentation \\via Random Fourier Feature-based Gaussian Process
}

\author{
\IEEEauthorblockN{
Issei Saito\IEEEauthorrefmark{1},
Masatoshi Nagano\IEEEauthorrefmark{1}\IEEEauthorrefmark{2}, 
Tomoaki Nakamura\IEEEauthorrefmark{1},
Daichi Mochihashi\IEEEauthorrefmark{3},
and Koki Mimura\IEEEauthorrefmark{4}
}
\IEEEauthorblockA{\IEEEauthorrefmark{1}The University of Electro-Communications, Tokyo, Japan \\
}
\IEEEauthorblockA{\IEEEauthorrefmark{2}Kyoto University, Kyoto, Japan}
\IEEEauthorblockA{\IEEEauthorrefmark{3}The Institute of Statistical Mathematics, Tokyo, Japan}
\IEEEauthorblockA{\IEEEauthorrefmark{4}National Center of Neurology and Psychiatry, Tokyo, Japan}
E-mail: i\_saito@radish.ee.uec.ac.jp}

\def\bq{\begin{equation}}
\def\eq{\end{equation}}
\def\beq{\begin{eqnarray}}
\def\eeq{\end{eqnarray}}
\def\ba{\begin{array}}
\def\ea{\end{array}}
\def\bc{\begin{center}}
\def\ec{\end{center}}

\def\dsum{\sum\limits}
\def\disp{\displaystyle}
\def\ejw{e^{j\omega}}
\def\ejwi{e^{j\omega_{i}}}
\def\e-jwi{e^{-j\omega_{i}}}
\def\dfrac#1#2{\disp{\frac{#1}{#2}}}
\def\teigi{\stackrel{\triangle}{=}}

\makeatletter
\def\lddots{\mathinner{\mkern1mu\raise\p@\vbox{\kern7\p@\hbox{.}}\mkern2mu
\raise4\p@\hbox{.}\mkern2mu\raise7\p@\hbox{.}\mkern1mu}}
\makeatother

\def\argmax{\mathop{\rm argmax}}

\def\baa{{ \boldsymbol a}}
\def\bb{{ \boldsymbol b}}
\def\bcc{{ \boldsymbol c}}
\def\bd{{ \boldsymbol d}}
\def\be{{ \boldsymbol e}}
\def\boldsymbolf{{ \boldsymbol f}}
\def\bg{{ \boldsymbol g}}
\def\bh{{ \boldsymbol h}}
\def\bi{{ \boldsymbol i}}
\def\bj{{ \boldsymbol j}}
\def\bk{{ \boldsymbol k}}
\def\bl{{ \boldsymbol l}}
\def\bm{{ \boldsymbol m}}
\def\bn{{ \boldsymbol n}}
\def\bo{{ \boldsymbol o}}
\def\bp{{ \boldsymbol p}}
\def\bqq{{ \boldsymbol q}}
\def\br{{ \boldsymbol r}}
\def\bs{{ \boldsymbol s}}
\def\bt{{ \boldsymbol t}}
\def\bu{{ \boldsymbol u}}
\def\bv{{ \boldsymbol v}}
\def\bw{{ \boldsymbol w}}
\def\bx{{ \boldsymbol x}}
\def\by{{ \boldsymbol y}}
\def\bz{{ \boldsymbol z}}

\def\bA{{ \boldsymbol A}}
\def\bB{{ \boldsymbol B}}
\def\bC{{ \boldsymbol C}}
\def\bD{{ \boldsymbol D}}
\def\bE{{ \boldsymbol E}}
\def\bF{{ \boldsymbol F}}
\def\bG{{ \boldsymbol G}}
\def\bH{{ \boldsymbol H}}
\def\bI{{ \boldsymbol I}}
\def\bJ{{ \boldsymbol J}}
\def\bK{{ \boldsymbol K}}
\def\bL{{ \boldsymbol L}}
\def\bM{{ \boldsymbol M}}
\def\bN{{ \boldsymbol N}}
\def\bO{{ \boldsymbol O}}
\def\bP{{ \boldsymbol P}}
\def\bQ{{ \boldsymbol Q}}
\def\bR{{ \boldsymbol R}}
\def\bS{{ \boldsymbol S}}
\def\bT{{ \boldsymbol T}}
\def\bU{{ \boldsymbol U}}
\def\bV{{ \boldsymbol V}}
\def\bW{{ \boldsymbol W}}
\def\bX{{ \boldsymbol X}}
\def\bY{{ \boldsymbol Y}}
\def\bZ{{ \boldsymbol Z}}

\def\b0{{\boldsymbol 0}}
\def\bPhi{{\boldsymbol\Phi}}
\def\bomega{{\boldsymbol\omega}}
\def\bLambda{{\boldsymbol\Lambda}}
\def\blambda{{\boldsymbol\lambda}}
\def\bmu{{\boldsymbol\mu}}
\def\bsigma{{\boldsymbol\sigma}}
\def\bnu{{\boldsymbol\nu}}
\def\bepsilon{{\boldsymbol\epsilon}}
\def\bphi{{\boldsymbol\phi}}
\def\bSigma{{\boldsymbol\Sigma}}
\def\bPhi{{\boldsymbol\Phi}}
\def\balpha{{\boldsymbol\alpha}}
\def\bTheta{{\boldsymbol\Theta}}
\def\btheta{{\boldsymbol\theta}}
\def\bGamma{{\boldsymbol\Gamma}}
\def\bPsi{{\boldsymbol\Psi}}
\def\bDelta{{\boldsymbol\Delta}}
\def\bPi{{\boldsymbol\Pi}}
\def\bXi{{\boldsymbol\Xi}}
\def\bxi{{\boldsymbol\xi}}
\def\bomega{{\boldsymbol\omega}}
\def\bOmega{{\boldsymbol\Omega}}

\def\balpha{{\boldsymbol\alpha}}
\def\bbeta{{\boldsymbol\beta}}
\maketitle

\begin{abstract}
In this paper, we propose RFF-GP-HSMM, a fast unsupervised time-series segmentation method that incorporates random Fourier features (RFF) to address the high computational cost of the Gaussian process hidden semi-Markov model (GP-HSMM). GP-HSMM models time-series data using Gaussian processes, requiring inversion of an $N \times N$ kernel matrix during training, where $N$ is the number of data points. As the scale of the data increases, matrix inversion incurs a significant computational cost. To address this, the proposed method approximates the Gaussian process with linear regression using RFF, preserving expressive power while eliminating the need for inversion of the kernel matrix. Experiments on the Carnegie Mellon University (CMU) motion-capture dataset demonstrate that the proposed method achieves segmentation performance comparable to that of conventional methods, with approximately 278 times faster segmentation on time-series data comprising 39,200 frames.

\end{abstract}
\begin{IEEEkeywords}
computational efficiency, Gaussian process, motion analysis, random Fourier features, time-series modeling, unsupervised segmentation

\end{IEEEkeywords}

\section{Introduction}
Unsupervised segmentation automatically extracts recurring patterns from time-series data.  
Segmentation enables the extraction of frequent events and facilitates analysis of their underlying structure, even in complex time-series data that are difficult to analyze manually.  
Therefore, it plays an important role in many real-world time-series processing tasks, including industrial applications.  
Furthermore, humans learn and recognize speech and motion by segmenting continuously perceived information into meaningful units, referred to as segments in this paper.
Reproducing this segmentation ability is important for understanding human cognitive functions and for enabling intelligence in robots, such as language comprehension and motion imitation.

Various approaches have been proposed for unsupervised segmentation \cite{fox2011joint, matsubara2014autoplait, gmmunsupervised, bojanowski2014weakly, huang2016connectionist, richard2017weakly}.  
Segmenting time-series data requires estimating the start and end points of each segment and classifying the segments into categories, which involves exploring all possible combinations of these parameters.  
As this search incurs extremely high computational costs, many studies have reduced the burden by introducing heuristic assumptions to determine the segmentation process \cite{bojanowski2014weakly, huang2016connectionist, richard2017weakly}.  
However, segmentation accuracy in such heuristic methods is often highly dependent on the specific characteristics of the task.

To address these issues, the Gaussian process hidden semi-Markov model (GP-HSMM) was proposed as a probabilistic generative model for segmentation \cite{nakamura2017segmenting}, achieving high segmentation accuracy in an unsupervised manner.  
The GP-HSMM uses Gaussian processes for modeling emission probability distributions, with each discretized state representing a pattern that recurs within the time-series data.  
In this model, segmentation is performed without labeled data by inferring the segmentation points and segments' classes from observed time-series data via Bayesian inference.  
The GP-HSMM has been successfully applied across various domains, including work analysis in industrial settings \cite{10311638}, behavioral analysis of animals \cite{mimura2024unsupervised}, studies on the emergence of language from continuous speech \cite{10644981}, and learning of robotic manipulation motions \cite{Mo03052023}.

However, despite its accuracy, the GP-HSMM suffers from high computational costs, particularly due to the inversion of the kernel matrix in the Gaussian process, resulting in long computation times.  
This limitation is particularly problematic in industrial applications, where real-time processing and the ability to handle large-scale data are essential for practical deployment.  
At present, no effective solution has been established to address the computational inefficiencies of GP-HSMM.  
To address this challenge, we propose RFF-GP-HSMM, a method designed to eliminate the computational bottleneck by significantly reducing the cost associated with Gaussian processes.  
The proposed method improves computational efficiency while minimizing any degradation in segmentation performance by using random Fourier features (RFF) \cite{NIPS2007_013a006f}, which approximate kernel functions using basis functions derived through the Fourier transformation, thereby avoiding costly matrix inversion.

In our experiments, we applied both the conventional GP-HSMM and the proposed RFF-GP-HSMM to motion-capture data of subjects performing exercises. The results confirmed that the proposed method substantially reduced computation time while maintaining segmentation accuracy.  
In particular, for a large-scale dataset comprising 39,200 frames (approximately 2 h and 43 min of time-series data), the proposed method achieved a remarkable improvement in computational efficiency, performing approximately 278 times faster than the conventional method.

The study enables real-time, scalable segmentation of large time-series data for practical applications in robotics, industry, and behavioral analysis.
\section{Related work}
Segmentation methods using supervised learning can capture patterns in time-series data with high accuracy and have been effectively used to analyze human and animal behaviors\cite{9067967, lea2017temporal, yeung2016end, diba2018spatio}.  
However, these methods require pretraining with a large volume of labeled data.  
Therefore, segmenting diverse time-series datasets requires significant time and effort for data annotation, posing a major constraint for applications that demand rapid analysis in real-world environments.  
Moreover, these methods are unsuitable for segmenting complex time-series data that is difficult to annotate.

To address this issue, segmentation methods based on semi-supervised learning, requiring only limited labeled data, have been proposed \cite{bojanowski2014weakly, huang2016connectionist, richard2017weakly}.  
These methods reduce the dependency on extensive training data compared to supervised learning methods.
However, they require prior specification of the order in which specific patterns appear, and the segmentation is learned based on this prior information.  
Therefore, such methods are effective only when time-series data exhibit fixed pattern sequences in a predetermined order.
For more complex time-series data, where the order of pattern appearance frequently varies, their effectiveness diminishes.
Hence, semi-supervised learning methods often require retraining whenever the pattern order changes, severely limiting their applicability.


In addition, an action-segmentation transformer \cite{liu2023temporal}, which applies the transformer architecture to time-series data to sequentially generate action labels, has been proposed.  
In the domains of sensor data and continuous signal analysis, models such as transformers for time series (TST) \cite{wen2022transformers} and GPT-TS \cite{das2023long} have also been introduced. These models can effectively capture long-term dependencies, identify change points, and model semantic transitions.  
Although such models exhibit high expressive power, they demand substantial computational resources, limiting their applicability in real-time processing or embedded environments.  
Moreover, they require large-scale datasets and extensive hyperparameter tuning, while their complex architectures pose challenges in interpretability and entail high implementation costs.

To address these problems, unsupervised learning methods for analyzing time-series data have been proposed \cite{gmmunsupervised, fox2011joint, matsubara2014autoplait}.  
These methods do not require pretraining and offer greater interpretability of the inference results.  
Sener and Yao demonstrated that videos can be segmented with high accuracy in an unsupervised manner by clustering using Gaussian mixture models (GMMs)\cite{gmmunsupervised}.  
However, this approach assumes that each action appears only once, rendering application to real-world time-series data that includes repeated actions difficult.

To probabilistically segment actions, models based on hidden Markov models (HMMs) have also been proposed \cite{fox2011joint, matsubara2014autoplait}.  
Fox et al. proposed a method that segments skeletal motion-capture data using HMMs in an unsupervised manner \cite{fox2011joint}, where contiguous data points classified into the same state are treated as segments.  
However, HMMs tend to produce frequent state transitions, resulting in overly fine segmentation.  
To overcome the limitations of single-HMM approaches, Matsubara et al. proposed AutoPlait\cite{matsubara2014autoplait}, which segments time-series data using multiple HMMs, each representing a recurring pattern, and identifies segments at the switching points between HMMs.  
Nonetheless, as HMMs represent time-series data using only the mean and variance, they are insufficient for capturing complex motion sequences.

To address this problem, we previously proposed GP-HSMM \cite{nakamura2017segmenting}, which uses Gaussian processes as a nonparametric model to capture recurring complex patterns and HSMMs to model the duration of those patterns.  
This method, based on a probabilistic generative model, offers higher interpretability and enables accurate segmentation of more complex time-series data than conventional methods.  
Hence, GP-HSMM has been widely applied in various studies owing to its balance of interpretability and segmentation accuracy.  
For example, Mimura et al. used a hierarchical Dirichlet process within the GP-HSMM framework\cite{8594029} to demonstrate that characteristic behaviors of marmosets can be extracted in an unsupervised manner \cite{mimura2024unsupervised}.  
In manufacturing environments, GP-HSMM has been used to analyze worker behavior, revealing not only work patterns but also operator-specific traits and proficiency levels \cite{10311638}.  
In the field of robot learning, the method has been utilized to segment continuous tasks into primitive actions \cite{Mo03052023}.
Despite its effectiveness in high-accuracy unsupervised segmentation, the GP-HSMM suffers from a significant computational burden owing to the Gaussian process component.  

A promising remedy is to derive a basis function that approximates the kernel using random Fourier features (RFFs) \cite{NIPS2007_013a006f}.
RFFs have enabled scalable GPLVMs \cite{zhang2023bayesian,li2024preventing} and accelerated hybrid HMMs without loss of accuracy \cite{jung2020scalable}.
Motivated by these results, we introduce RFF into GP-HSMM and propose the computationally efficient RFF-GP-HSMM.

\section{RFF-GP-HSMM}
\subsection{Generative process}
Fig. \ref{fig:model} shows a graphical model of the proposed RFF-GP-HSMM.  
This model assumes that time-series data are generated through the following process:

\begin{figure}[t]
  \begin{center}
    \includegraphics[scale=0.8]{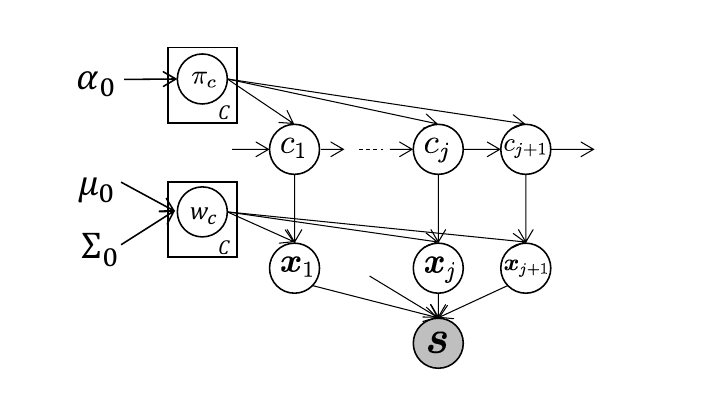}
    \caption{Graphical model of the RFF-GP-HSMM.}
    \label{fig:model}
  \end{center}
  \vspace{-0.4cm}
\end{figure}

First, the transition probabilities are generated from a Dirichlet distribution parameterized by $\alpha_0$:
\begin{eqnarray}
\pi_c &\sim& P(\pi | \alpha_0 ).
\end{eqnarray}
The $j$th segment class $c_j$ is generated using the $j-1$th class $c_{j-1}$ and the transition probability $\pi_c$:
\begin{eqnarray}
c_j &\sim& P(c | c_{j-1}, \pi_c).
\end{eqnarray}

In conventional GP-HSMM, each segment $\bx_j$ is assumed to be generated by a Gaussian process.
By contrast, RFF-GP-HSMM assumes that each segment is generated via linear regression.  
The regression weights $\bw$ are generated from a Gaussian distribution with mean $\bmu_0$ and covariance matrix $\bSigma_0$:
\begin{eqnarray}
\bw_c &\sim&  {\mathcal N}(\bw|\bmu_0, \bSigma_0).
\end{eqnarray}
Segment $\bx_j$, corresponding to class $c_j$, is generated using linear regression with weights $\bw_{c}$:
\begin{eqnarray}
\bx_j &\sim& p(\bx | \bw_{c_j}).
\end{eqnarray}

The observed time-series data $\bS$ are generated by concatenating the segments $\bx_j$.  
However, simple linear regression alone cannot adequately capture complex time-series patterns.  
To address this, RFF-GP-HSMM uses linear regression based on basis functions derived from RFF\cite{NIPS2007_013a006f}.


\subsection{Gaussian process approximation using RFF}
\subsubsection{Gaussian processes}
In conventional GP-HSMM, complex and varying time-series data are modeled using Gaussian processes to predict the output value $x'$ at time $t'$ for each class.  
A Gaussian process is a probabilistic model that applies the marginalization of weights $\bw$ and utilizes the kernel trick within a linear regression model $x' = \bw^T \bphi(t')$, where  $\bphi$ represents the basis functions.

Given a set of segments $\bX_c$ classified into class $c$ and the corresponding set of time steps $\bt_c$, the predictive distribution of the output $x'$ at time $t'$ in a Gaussian process is given by the following Gaussian distribution:
\begin{equation}
p(x'|t',\bX_c, \bt_c)  = {\mathcal N}(\bk^{T}\bK^{-1}\bX_c, k(\hat{t'}, \hat{t'})-\bk^{T}\bK^{-1}\bk), \label{eq:gp_pred}
\end{equation}
where $k(\cdot, \cdot)$ is a kernel function, and $\bK$ is a matrix whose $p$-row $q$-column elements are given by the following expression:
\begin{equation}
\label{equ:covariance_func}
K(p, q)=k(t_{p},t_{q})+\beta^{-1}\delta_{pq}, 
\end{equation}
where $t_p$ and $t_q$ are the $p$-th and $q$-th elements of $\bt_c$, respectively.  
$\beta$ represents the noise variance, and $\delta_{pq}$ is the Kronecker delta function.  
$\bk$ is a vector whose $p$-th element is $k(t_p, t')$.  
In this paper, the following radial basis function (RBF) kernel was used:
\begin{equation}
k(t_p,t_q)=\exp(-\frac{1}{2}||t_{p}-t_{q}||^{2}), \label{eq:rbf}
\end{equation}
By using the kernel, input--output relationships in a nonlinear high-dimensional space can be flexibly captured without explicitly computing the basis function $\bphi(\cdot)$, but by computing the inner product $k(t_p, t_q) = \bphi(t_p)^T \bphi(t_q)$.

In Gaussian processes, computing (\ref{eq:gp_pred}) involves inverting the matrix $\bK$.  
If the number of data points in $\bX_c$ is denoted by $N_c$, then the matrix $\bK$ is of size $N_c \times N_c$.  
Thus, as the number of time-series data points $N_c$ increases, the size of $\bK$ also increases, and the matrix inversion becomes increasingly computationally expensive.  
This poses a significant computational bottleneck in conventional GP-HSMM.

\subsubsection{Random Fourier feature}
In Gaussian processes, the input--output relationships are captured in a nonlinear high-dimensional space without explicitly computing the basis function $\bphi$, owing to the kernel trick.  
However, if there exists a basis function $\bphi$ such that $k(t_p, t_q) \approx \bphi(t_p)^T \bphi(t_q)$ approximates the RBF kernel via inner products, a linear regression model $x' = \bw^T \bphi(t')$ can be applied without the need to compute $\bK$.  
A method for deriving such a basis function that approximates a given kernel is known as the RFF method~\cite{NIPS2007_013a006f}.

In RFF, the Fourier transform of the kernel function is first considered.  
The RBF kernel in (\ref{eq:rbf}) is rewritten as a function $k'(\Delta) = \exp(-\|\Delta\|^2 / 2)$ using the input difference $\Delta = t_p - t_q$, and its Fourier transform becomes the following Gaussian distribution:
\begin{equation}
p(\omega) = (2 \pi)^{-\frac{D}{2}} \exp(-\frac{1}{2}||\omega||^2). 
\end{equation}
Next, the original kernel function is recovered by applying an inverse Fourier transform to $p(\omega)$.
\begin{eqnarray}
k(t_p, t_q) &=& k'(\Delta) \\
&=& \int_\mathbb{R} p(\omega) \exp( i \omega (t_p - t_q) ) d \omega \\
&=& \int_\mathbb{R} p(\omega)\cos(\omega (t_p - t_q)) + i \sin(\omega (t_p - t_q)) d \omega.  \nonumber \\
\end{eqnarray}
This transformation uses Euler's formula $\exp(iz) = \cos(z) + i \sin(z)$.  
Furthermore, as the output of the kernel function is a real value, we obtain:
\begin{equation}
k(t_p, t_q) = \int_\mathbb{R} p(\omega)\cos(\omega (t_p - t_q)) \, d \omega 
\end{equation}
Next, by introducing a random variable $b \sim \mathrm{Uniform}(0, 2\pi)$ (with $p(b) = \frac{1}{2\pi}$) and performing marginalization over $b$, we obtain:
\begin{eqnarray}
&&\int_0^{2 \pi} p(b) p(\omega)\cos(\omega (t_p - t_q)) d b = p(\omega)\cos(\omega (t_p - t_q)), \nonumber \\ \\
&&\int_0^{2 \pi} p(b) p(\omega)\cos(\omega (t_p + t_q) + 2 b) d b \nonumber \\
&&~~~~~~~~~~~= \left[ \frac{1}{4 \pi} p(\omega)\sin(\omega (t_p + t_q) + 2 b)  \right]_0^{2 \pi}=0.
\end{eqnarray}
Hence, $k(t_p, t_q)$ can be rewritten as follows:
\begin{eqnarray}
k(t_p, t_q) = \int_\mathbb{R} \int_0^{2 \pi} p(b) p(\omega) \left\{ \cos(\omega (t_p + t_q) + 2 b) \right. \nonumber \\
~~~ \left. +\cos(\omega (t_p - t_q)) \right\} db d \omega. 
\end{eqnarray}
Let $A = \omega t_p + b$ and $B = \omega t_q + b$. Using the following equation:
\begin{eqnarray}
\cos(A + B) + \cos(A - B) = 2 \cos(A)\cos(B),
\end{eqnarray}
we can further simplify the expression as follows:
\begin{eqnarray}
k(t_p, t_q) &=& \int_\mathbb{R} \int_0^{2 \pi} p(b) p(\omega) \sqrt{2} \cos(\omega t_p +b) \nonumber \\
&&~~~~~~ \times \sqrt{2} \cos(\omega t_q +b) db d \omega 
\end{eqnarray}
This integral computation is approximated using Monte Carlo estimation with $M$ samples $\omega^{(m)}$ and $b^{(m)}$:
\begin{eqnarray}
\omega^{(m)} &\sim& p(\omega), \\
b^{(m)} &\sim& {\rm Uniform}(0, 2 \pi), \\
k(x, y) &\approx& \frac{1}{M} \sum_m  \sqrt{2}\cos(\omega^{(m)} t_p +b^{(m)}) \nonumber \\
&&~~~~~~~~ \times \sqrt{2}\cos(\omega^{(m)} t_q +b^{(m)}).  \label{eq:rff_mc}
\end{eqnarray}
Equation (\ref{eq:rff_mc}) can be interpreted as the inner product $k(t_p, t_q) = \bphi(t_p)^T \bphi(t_q)$ of $M$-dimensional vectors, onto which $t_p$ and $t_q$ are projected using the basis function:
\begin{eqnarray}
\bphi(t) =  
\begin{bmatrix}
\sqrt{\frac{2}{M}} \cos(\omega^{(1)} t +b^{(1)}) \\
\sqrt{\frac{2}{M}} \cos(\omega^{(2)} t +b^{(2)}) \\
\vdots \\
\sqrt{\frac{2}{M}} \cos(\omega^{(M)} t +b^{(M)}) \\
\end{bmatrix}.
\end{eqnarray}
In other words, $\bphi(t)$ serves as a basis function that approximates a Gaussian kernel.

\subsubsection{Efficient Gaussian process approximation using RFF}
Using RFF, a basis function capable of constructing an RBF kernel is explicitly derived.  
Therefore, the Gaussian process regression in (\ref{eq:gp_pred}) is approximated by linear regression using this basis function.  
In Bayesian linear regression, given a set of segments $\bX_c$ classified into class $c$ and the corresponding set of time steps $\bt_c$, the predictive distribution of the output $x'$ at time $t'$ is given by the following Gaussian distribution:
\begin{eqnarray}
p(x'|t', \bX_c, \bt_c ) &=& {\mathcal N}(x'|\mu_c(t'), \sigma^2_c(t')), \\
\mu_c(t') &=& \bm_c^T \bphi(t'), \\
\sigma^2_c(t') &=& \beta^{-1} + \bphi(t')^T \bSigma_c \bphi(t'), \\
\bSigma_c &=& \left( \psi \bI +\beta \sum_{p=1}^{N_c} \bphi(t_p) \bphi(t_p)^T \right )^{-1}, \label{eq:lin_reg_inv}\\
\bm_c &=& \beta \bS \sum_{p=1}^{N_c} x_p \bphi(t_p).
\end{eqnarray}
Here, $\bI$ denotes the $M \times M$ identity matrix, and the prior distribution of weights $\bw$ is assumed to follow a Gaussian distribution with mean $\bmu_0 = \b0$ and covariance matrix $\bSigma_0 = \psi \bI$.  
$\beta$ is a parameter representing the observation noise, and $x_p$ and $t_p$ denote the $p$-th elements of $\bX_c$ and $\bt_c$, respectively.

While the predictive distribution of Gaussian processes in  (\ref{eq:gp_pred}) requires the inversion of an $N_c \times N_c$ matrix $\bK$, the predictive distribution using RFF can be computed by inverting an $M \times M$ matrix in (\ref{eq:lin_reg_inv}).  
As $N_c$ increases with the data size, the computational cost of inverting $\bK$ grows accordingly.  
In contrast, $M$ is a fixed hyperparameter; thus, the computational cost of matrix inversion in RFF-based linear regression remains constant regardless of the number of data points.

In RFF-GP-HSMM, when the output is a multidimensional vector $\bx'$, each dimension is assumed to be generated independently, and the generation probability is calculated as follows:
\begin{equation}
\label{equ:multi_dim}
p(\bx'|t', \bX_{c}, \bt_c) = \prod_d^D p(\bx'^{(d)}|t', \bX^{(d)}_{c}, \bt_c),
\end{equation}
where the superscript $d$ indicates the $d$-th dimension of each variable.

\subsection{Parameter inference}
\subsubsection{Blocked Gibbs sampler}
Segments and their corresponding classes in the observation sequence were inferred using a blocked Gibbs sampler.  
The inference procedure for the blocked Gibbs sampler is as follows:
\begin{enumerate}
 \item Randomly segment all observation sequences $\bs_1, \bs_2, \ldots, \bs_n, \ldots$, and initialize the model parameters.
 \item For the $n$-th observation sequence $\bs_n$, remove the segments $\bx_{nj}$ from their assigned classes, and update the parameters of linear regression for each class ($\bSigma_c, \bm_c$) and the transition probabilities $P(c \mid c')$.
 \item Sample new segments $\bx_{nj}$ ($j = 1, 2, \dots, J$) and their corresponding classes $c_{nj}$ ($j = 1, 2, \dots, J$) using the forward filtering–backward sampling procedure described later, and update the linear regression parameters $\bSigma_c, \bm_c$ and transition probabilities $P(c \mid c')$.
 \item Repeat steps 2 and 3 for all sequences until convergence.
\end{enumerate}
Using this procedure, the parameters of the RFF-GP-HSMM can be optimized, and the observation sequences can be segmented.


\subsubsection{Forward filtering-backward sampling}
Forward filtering–backward sampling (FFBS) was used to sample the segments and their corresponding classes.  
FFBS is a method that efficiently computes the probabilities of all possible segmentation and classification patterns using dynamic programming and samples segments and classes based on these probabilities.  
In the forward filtering step, we compute the forward probability $\alpha$, which represents the probability that a subsequence of length $k$ ending at time step $t$ in the observation sequence $\bs$ forms a segment belonging to class $c$:
\begin{eqnarray}
\label{equ:ffbs}
&& \alpha[t][k][c] =  \left( \prod_{t'=t-k}^t p(\bs_{n,t'} |t', \bX_{c}, \bt_c)\right) p(k|\lambda) \nonumber \\
&& ~~~~~~~~~~~~\times\sum_{k'=K_{\rm min}}^{K_{\rm max}}\sum_{c'=1}^{C} p(c|c')\alpha[t-k][k'][c'],
\end{eqnarray}
where $\bs_{n,t'}$ denotes the data point at time step $t'$ in the sequence $\bs_n$.  
$C$ is the number of segment classes, and $K_{\rm min}$ and $K_{\rm max}$ represent the minimum and maximum allowable segment lengths, respectively.  
The term $p(k | \lambda)$ is the probability of a segment having length $k$, which is modeled using a Poisson distribution with mean segment length $\lambda$:
\begin{equation}
\label{equ:avelen}
P(k|\lambda)=\frac{\lambda^{k}e^{-\lambda}}{k!}.
\end{equation}
$p(c \mid c')$ represents the transition probability between classes and is given by the following Dirichlet multinomial model:
\begin{equation}
\label{equ:trans_prob}
p(c|c') = \frac{N_{c'c}+\alpha}{N_{c'}+C\alpha}, 
\end{equation}
where $N_{c'}$ denotes the number of segments assigned to class $c'$, and $N_{c'c}$ denotes the number of transitions from class $c'$ to class $c$.  
$k'$ and $c'$ represent the length and class of the segment preceding $\bs_{t-k:t}$, respectively.
As these values are unknown at this stage, (\ref{equ:ffbs}) marginalizes over all possible lengths and classes.

Next, backward sampling is performed based on the forward probabilities.
In backward sampling, the class and segment lengths are sampled in reverse, starting from the end of the time-series data ($t = T$), according to the forward probabilities.

\begin{equation}
\label{equ:backward}
k, c \sim p( c_{j-1} | c ) \alpha[t][k][c].
\end{equation}
This procedure enables the determination of the segments and their corresponding classes in the observation sequence.  
The overall algorithm is summarized in Algorithm~\ref{alg:ffbs}.

\begin{algorithm}[t]
\small
\caption{Forward filtering-backward sampling.}
\label{alg:ffbs}
\begin{algorithmic}[1]
\STATE // Forward filtering
\FOR{ $t=1$ to $T$}
	\FOR{ $k=1$ to $K$}
		\FOR{ $c=1$ to $C$}
			\STATE Compute $\alpha[t][k][c]$
		\ENDFOR  
	\ENDFOR  
\ENDFOR  

\STATE 
\STATE // Backward sampling
\STATE $t = T, j=1$
\WHILE{ $t>0$ }
	\STATE $k, c \sim p( c_{j+1} | c ) \alpha[t][k][c]$
	\STATE $\bx_j = \bs_{t-k:t}$
	\STATE $c_j = c$
	\STATE $t = t-k$
	\STATE $j=j+1$
\ENDWHILE
\STATE return $(\bx_{J_n},\bx_{J_n-1} , \cdots ,\bx_1), (c_{J_n}, c_{J_n-1}, \cdots, c_1)$
\end{algorithmic}
\end{algorithm}
\section{Experiments}\label{result}
\subsection{Experimental setup}\label{setup}
\begin{figure}[t]
	\begin{center}
	\includegraphics[scale=0.4]{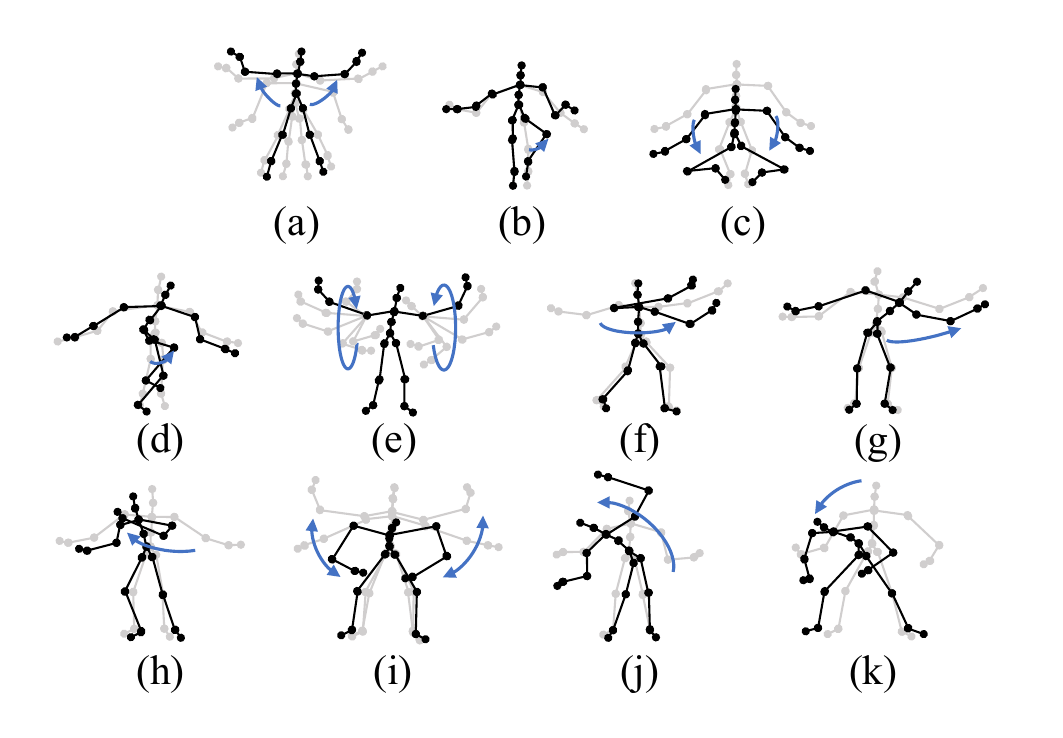}
	\caption{Motions included in exercise motion: (a) jumping
jack, (b) jogging, (c) squatting, (d) knee raise, (e) arm circle, (f) twist, (g) side reach, (h) boxing, (i) arm wave, (j) side bend, and (k) toe touch.}
	\label{fig:motion}
	\end{center}
	\vspace{-0.4cm}
\end{figure}

In this experiment, we used three sequences from Subject 14 in the Carnegie Mellon University (CMU) motion-capture dataset\footnote{http://mocap.cs.cmu.edu}.  
These sequences include various activities such as running and jumping, as shown in Fig. \ref{fig:motion}.
The data were downsampled to 4 fps.  
We used 8-dimensional coordinate data comprising the 2-dimensional positions of the left hand, right hand, left foot, and right foot, and normalized the data to the range $[-1, 1]$ using min-max normalization.

For the GP-HSMM, we used an implementation that incorporates the fast forward filtering method proposed by Sasaki et al.~\cite{SASAKI20239691}.  
The proposed RFF-GP-HSMM method extends this implementation by introducing RFF, with the number of random features set to $M = 20$.  
For both models, the segment length parameters were set to $\lambda=20$, $K_{\rm min}=15$, and $K_{\rm max}=30$. 
The number of iterations of the blocked Gibbs sampler was set to five.

The specifications of the PC used in this experiment are listed in Table~\ref{table:mac}.

\begin{table}[t]
\caption{Computer used in the experiments.}
\label{table:mac}
 \centering
\begin{tabular}{cc}
    \hline
    CPU & Apple M1 Ultra \\
    Memory & 64GB \\
    \hline
\end{tabular}
\end{table}

\subsection{Experiment 1: Segmentation accuracy}\label{result}

We evaluated the segmentation accuracy of the conventional GP-HSMM and the proposed RFF-GP-HSMM.  
The dataset, comprising the three sequences described in Section \ref{setup}, was duplicated 10 times, resulting in a total of 30 sequences.

As both GP-HSMM and RFF-GP-HSMM are sensitive to initial values, we used the result with the highest likelihood among the 10 trials for evaluation.  
As the evaluation metric, we used the normalized Hamming distance (NHD) between the ground truth labels and the segmentation results.  
The NHD is in the range $[0, 1]$, with values closer to $0$ indicating greater consistency between the segmentation results and the ground truth.

Table \ref{table:ari} shows the NHD values for each method.  
Both methods achieved an accuracy of approximately 0.35, demonstrating that the proposed RFF-based approximation method achieves segmentation accuracy comparable to that of the conventional GP-based method.

\begin{table}[t]
\caption{Normalized Hamming distance of segmentation.}
\centering
\begin{tabular}{cc}
    \hline
     & Normalized Hamming distance  \\
    \hline
    GP-HSMM & 0.351  \\
    RFF-GP-HSMM & 0.357 \\
     \hline
     \label{table:ari}
  \end{tabular}
  \end{table}

\subsection{Experiment 2: Computational time}\label{sec:perplexity}
Next, we evaluated the computation time for each method.  
The number of data points was increased by duplicating the three sequences used in Experiment 1; the computation time was measured for datasets comprising 3, 30, 60, 90, 120, 150, 180, 210, and 240 sequences.  
As the three-sequence dataset contained 490 frames, the number of frames corresponding to 30-240 sequences were 4900, 9800, 14,700, 19,600, 24,500, 29,400, 34,300, and 39,200 frames, respectively.  
The frame rate of the data used in this experiment was 4 fps; thus, 39,200 frames corresponded to approximately 2 hours and 43 minutes of sequence length.
Each method was trained five times with different initial values, and the average computation time across the five trials was calculated.

\begin{figure}[t]
	\begin{center}
	\includegraphics[scale=0.35]{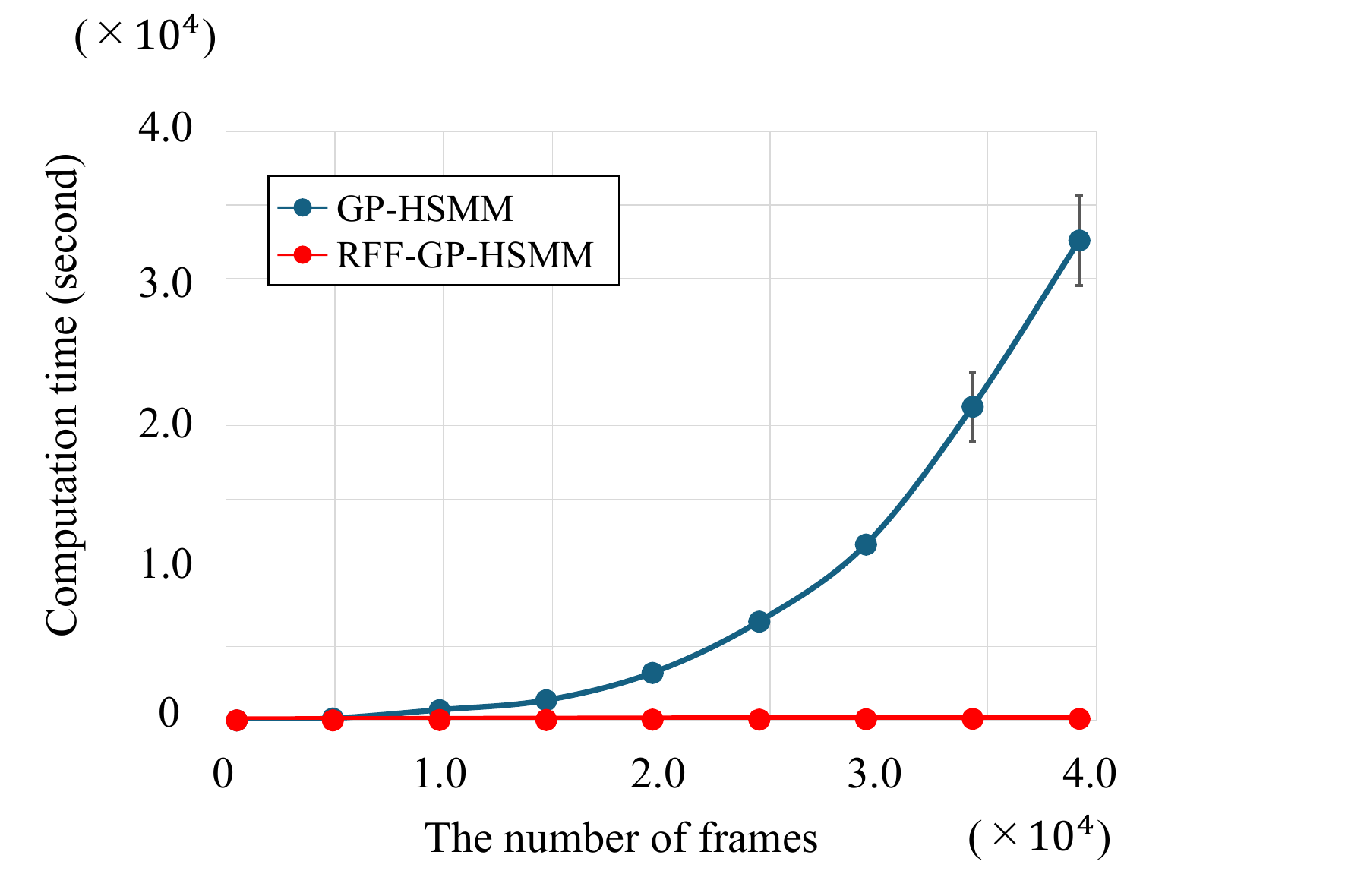}
	\caption{Computation time of GP-HSMM and RFF-GP-HSMM.}
	\label{fig:compare_time}
	\end{center}
	\vspace{-0.4cm}
\end{figure}

\begin{figure}[t]
	\begin{center}
	\includegraphics[scale=0.35]{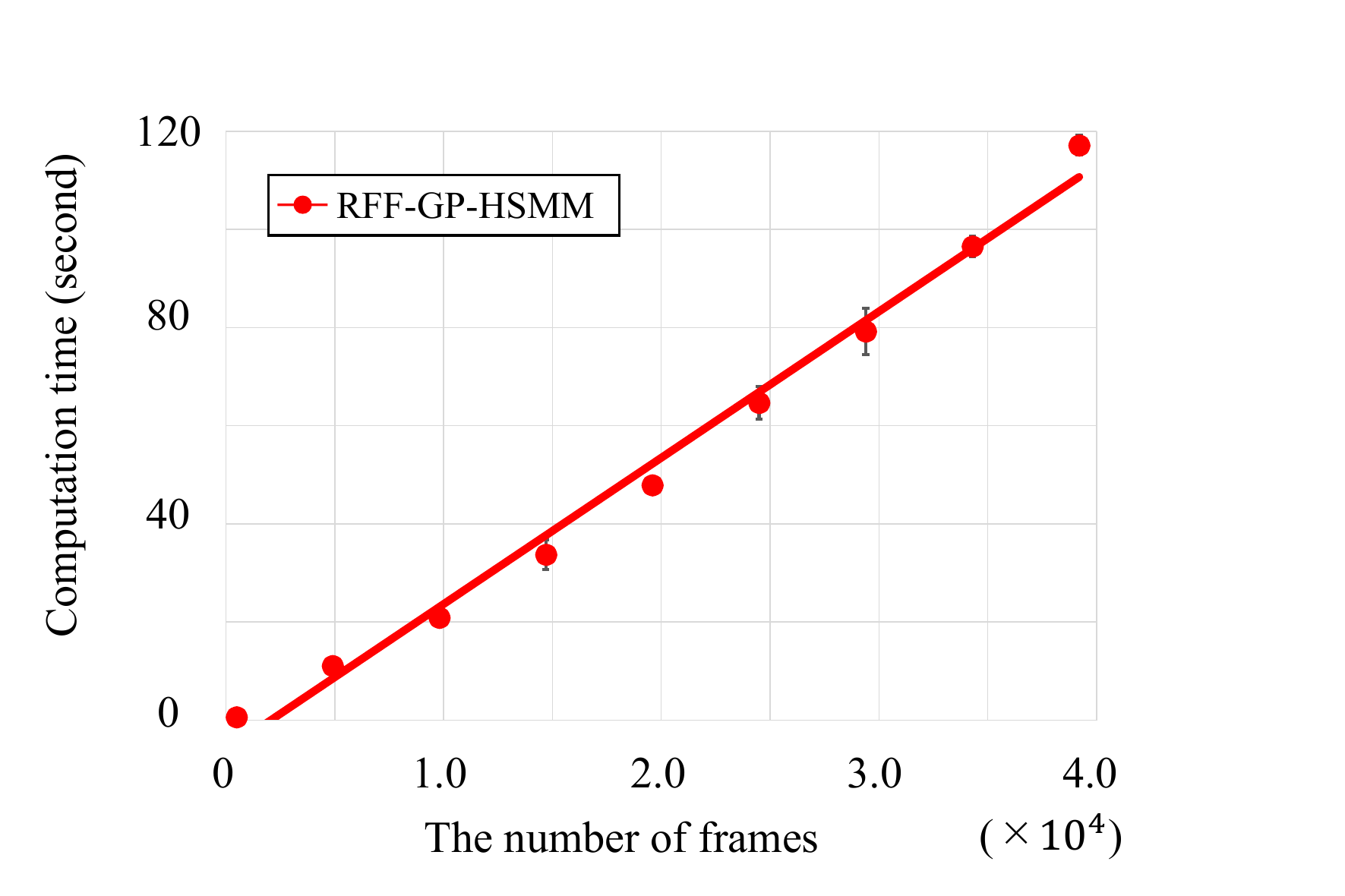}
	\caption{Computation time of RFF-GP-HSMM.}
	\label{fig:rff_time}
	\end{center}
	\vspace{-0.4cm}
\end{figure}

Fig.~\ref{fig:compare_time} shows the computation time for each number of frames for both GP-HSMM and RFF-GP-HSMM. Conversely, Fig.~\ref{fig:rff_time} highlights only the computation times for RFF-GP-HSMM extracted from Fig.~\ref{fig:compare_time}.  
The graphs clearly indicate that the proposed RFF-GP-HSMM significantly reduces computation time compared to the conventional GP-HSMM.  
For the dataset with 39,200 frames, GP-HSMM required $3.3 \times 10^4$ s (approximately 9 h), while RFF-GP-HSMM required only $1.2 \times 10^2$ s (approximately 2 min), achieving a speed-up of approximately 278 times.

In addition to the graphical comparison, Table~\ref{tab:time_speedup} summarizes the computation times and the corresponding speed-up ratios for all dataset sizes.  
This table clearly demonstrates that the proposed method consistently outperforms the conventional GP-HSMM as the dataset size increases.

\begin{table}[t]
\centering
\caption{
Computation times and speed-up ratio between GP-HSMM and RFF-GP-HSMM.
The speed-up ratio is defined as the ratio of the computation time of GP-HSMM to that of RFF-GP-HSMM.
}
\label{tab:time_speedup}
\begin{tabular}{cccc}
\hline
\textbf{Frames} & \textbf{GP-HSMM (s)} & \textbf{RFF-GP-HSMM (s)} & \textbf{Speed-up ratio} \\
	& $t_{\rm gp}$ & $t_{\rm rff}$ & $t_{\rm gp}/t_{\rm rff}$ \\
\hline
490   & 1    & 1    & 1 \\
4900  & 133  & 11   & 12 \\
9800  & 708  & 21   & 34 \\
14700 & 1361 & 34   & 40 \\
19600 & 3227 & 48   & 67 \\
24500 & 6713 & 65   & 103 \\
29400 & 11939 & 79  & 151 \\
34300 & 21294 & 97  & 220 \\
39200 & 32583 & 117 & 278 \\
\hline
\end{tabular}
\end{table}

Sasaki et al.~\cite{SASAKI20239691} reported that optimizing Gaussian process computations accelerated the conventional GP-HSMM by approximately four times on a dataset of approximately 8,500 frames.  
In contrast, the proposed method achieved a 34-fold speed-up on a comparable dataset with 9,800 frames, demonstrating that RFF-GP-HSMM enables extremely fast computation.

\section{Conclusion}
In this paper, we proposed RFF-GP-HSMM, which approximates Gaussian processes using random Fourier features (RFF) to address the high computational cost of the Gaussian process hidden semi-Markov model (GP-HSMM).  
The proposed method successfully reduced computation time by eliminating the need for kernel matrix inversion while maintaining the high segmentation accuracy of the conventional GP-HSMM.  
Experiments using the CMU motion-capture dataset demonstrated that the proposed method achieved segmentation accuracy comparable to that of the conventional method while significantly reducing computation time for large-scale data.  
In particular, it enabled fast segmentation of long and large-scale time-series data, which had been challenging for conventional methods, thereby improving real-time performance and scalability.  
These results suggest that the proposed method has broad applicability to various domains requiring time-series data segmentation, such as motion and speech, and is expected to contribute to practical applications in areas such as robotic control, industrial task analysis, and animal behavior understanding.

Future work includes evaluating the proposed method on a wider range of datasets, such as large-scale speech and industrial time-series data, to further assess its generality. We also plan to compare it with acceleration techniques based on auxiliary variables to clarify trade-offs between accuracy and computational efficiency, with the goal of enhancing its applicability to industrial use and practical time-series data analysis.

The implementation of RFF-GP-HSMM used in this study will be made available on GitHub\footnote{https://github.com/naka-lab/RFF-GP-HSMM}.

\section*{Acknowledgment}
This research was supported by AMED under Grant Number JP24wm0625124.

\bibliography{IEEEfull.bib}

\begin{thebibliography}{10}
\providecommand{\url}[1]{#1}
\csname url@samestyle\endcsname
\providecommand{\newblock}{\relax}
\providecommand{\bibinfo}[2]{#2}
\providecommand{\BIBentrySTDinterwordspacing}{\spaceskip=0pt\relax}
\providecommand{\BIBentryALTinterwordstretchfactor}{4}
\providecommand{\BIBentryALTinterwordspacing}{\spaceskip=\fontdimen2\font plus
\BIBentryALTinterwordstretchfactor\fontdimen3\font minus
  \fontdimen4\font\relax}
\providecommand{\BIBforeignlanguage}[2]{{%
\expandafter\ifx\csname l@#1\endcsname\relax
\typeout{** WARNING: IEEEtran.bst: No hyphenation pattern has been}%
\typeout{** loaded for the language `#1'. Using the pattern for}%
\typeout{** the default language instead.}%
\else
\language=\csname l@#1\endcsname
\fi
#2}}
\providecommand{\BIBdecl}{\relax}
\BIBdecl

\bibitem{fox2011joint}
E.~B. Fox, E.~B. Sudderth, M.~I. Jordan, and A.~S. Willsky, ``Joint modeling of
  multiple related time series via the beta process,'' 2011.

\bibitem{matsubara2014autoplait}
Y.~Matsubara, Y.~Sakurai, and C.~Faloutsos, ``Autoplait: Automatic mining of
  co-evolving time sequences,'' in \emph{Proceedings of the 2014 ACM SIGMOD
  international conference on Management of data}, 2014, pp. 193--204.

\bibitem{gmmunsupervised}
F.~Sener and A.~Yao, ``Unsupervised learning and segmentation of complex
  activities from video,'' in \emph{Proceedings of the IEEE Conference on
  Computer Vision and Pattern Recognition}, 2018, pp. 8368--8376.

\bibitem{bojanowski2014weakly}
P.~Bojanowski, R.~Lajugie, F.~Bach, I.~Laptev, J.~Ponce, C.~Schmid, and
  J.~Sivic, ``Weakly supervised action labeling in videos under ordering
  constraints,'' in \emph{Computer Vision--ECCV 2014: 13th European Conference,
  Zurich, Switzerland, September 6-12, 2014, Proceedings, Part V 13}.\hskip 1em
  plus 0.5em minus 0.4em\relax Springer, 2014, pp. 628--643.

\bibitem{huang2016connectionist}
D.-A. Huang, L.~Fei-Fei, and J.~C. Niebles, ``Connectionist temporal modeling
  for weakly supervised action labeling,'' in \emph{Computer Vision--ECCV 2016:
  14th European Conference, Amsterdam, The Netherlands, October 11--14, 2016,
  Proceedings, Part IV 14}.\hskip 1em plus 0.5em minus 0.4em\relax Springer,
  2016, pp. 137--153.

\bibitem{richard2017weakly}
A.~Richard, H.~Kuehne, and J.~Gall, ``Weakly supervised action learning with
  rnn based fine-to-coarse modeling,'' in \emph{Proceedings of the IEEE
  conference on Computer Vision and Pattern Recognition}, 2017, pp. 754--763.

\bibitem{nakamura2017segmenting}
T.~Nakamura, T.~Nagai, D.~Mochihashi, I.~Kobayashi, H.~Asoh, and M.~Kaneko,
  ``Segmenting continuous motions with hidden semi-markov models and gaussian
  processes,'' \emph{Frontiers in neurorobotics}, vol.~11, p.~67, 2017.

\bibitem{10311638}
I.~Saito, T.~Nakamura, T.~Hatta, W.~Fujita, S.~Watanabe, and S.~Miwa,
  ``Unsupervised work behavior analysis using hierarchical probabilistic
  segmentation,'' in \emph{IECON 2023- 49th Annual Conference of the IEEE
  Industrial Electronics Society}, 2023, pp. 1--6.

\bibitem{mimura2024unsupervised}
K.~Mimura, J.~Matsumoto, D.~Mochihashi, T.~Nakamura, H.~Nishijo, M.~Higuchi,
  T.~Hirabayashi, and T.~Minamimoto, ``Unsupervised decomposition of natural
  monkey behavior into a sequence of motion motifs,'' \emph{Communications
  Biology}, vol.~7, no.~1, p. 1080, 2024.

\bibitem{10644981}
I.~Saito, T.~Nakamura, A.~Taniguchi, T.~Taniguchi, Y.~Hayamizu, and S.~Zhang,
  ``Emergence of continuous signals as shared symbols through emergent
  communication,'' in \emph{2024 IEEE International Conference on Development
  and Learning (ICDL)}, 2024, pp. 1--6.

\bibitem{Mo03052023}
\BIBentryALTinterwordspacing
Y.~Mo, H.~Sasaki, T.~Matsubara, and K.~Yamazaki, ``Multi-step motion learning
  by combining learning-from-demonstration and policy-search,'' \emph{Advanced
  Robotics}, vol.~37, no.~9, pp. 560--575, 2023. [Online]. Available:
  \url{https://doi.org/10.1080/01691864.2022.2163187}
\BIBentrySTDinterwordspacing

\bibitem{NIPS2007_013a006f}
A.~Rahimi and B.~Recht, ``Random features for large-scale kernel machines,'' in
  \emph{Advances in Neural Information Processing Systems}, J.~Platt,
  D.~Koller, Y.~Singer, and S.~Roweis, Eds., vol.~20.\hskip 1em plus 0.5em
  minus 0.4em\relax Curran Associates, Inc., 2007.

\bibitem{9067967}
T.~Kobayashi, Y.~Aoki, S.~Shimizu, K.~Kusano, and S.~Okumura, ``Fine-grained
  action recognition in assembly work scenes by drawing attention to the
  hands,'' in \emph{2019 15th International Conference on Signal-Image
  Technology Internet-Based Systems (SITIS)}, 2019, pp. 440--446.

\bibitem{lea2017temporal}
C.~Lea, M.~D. Flynn, R.~Vidal, A.~Reiter, and G.~D. Hager, ``Temporal
  convolutional networks for action segmentation and detection,'' in
  \emph{proceedings of the IEEE Conference on Computer Vision and Pattern
  Recognition}, 2017, pp. 156--165.

\bibitem{yeung2016end}
S.~Yeung, O.~Russakovsky, G.~Mori, and L.~Fei-Fei, ``End-to-end learning of
  action detection from frame glimpses in videos,'' in \emph{Proceedings of the
  IEEE conference on computer vision and pattern recognition}, 2016, pp.
  2678--2687.

\bibitem{diba2018spatio}
A.~Diba, M.~Fayyaz, V.~Sharma, M.~M. Arzani, R.~Yousefzadeh, J.~Gall, and
  L.~Van~Gool, ``Spatio-temporal channel correlation networks for action
  classification,'' in \emph{Proceedings of the European Conference on Computer
  Vision (ECCV)}, 2018, pp. 284--299.

\bibitem{liu2023temporal}
Z.~Liu, L.~Wang, D.~Zhou, J.~Wang, S.~Zhang, Y.~Bai, E.~Ding, and R.~Fan,
  ``Temporal segment transformer for action segmentation,'' \emph{arXiv
  preprint arXiv:2302.13074}, 2023.

\bibitem{wen2022transformers}
Q.~Wen, T.~Zhou, C.~Zhang, W.~Chen, Z.~Ma, J.~Yan, and L.~Sun, ``Transformers
  in time series: A survey,'' \emph{arXiv preprint arXiv:2202.07125}, 2022.

\bibitem{das2023long}
A.~Das, W.~Kong, A.~Leach, S.~Mathur, R.~Sen, and R.~Yu, ``Long-term
  forecasting with tide: Time-series dense encoder,'' \emph{arXiv preprint
  arXiv:2304.08424}, 2023.

\bibitem{8594029}
M.~Nagano, T.~Nakamura, T.~Nagai, D.~Mochihashi, I.~Kobayashi, and M.~Kaneko,
  ``Sequence pattern extraction by segmenting time series data using gp-hsmm
  with hierarchical dirichlet process,'' in \emph{2018 IEEE/RSJ International
  Conference on Intelligent Robots and Systems (IROS)}, 2018, pp. 4067--4074.

\bibitem{zhang2023bayesian}
M.~M. Zhang, G.~W. Gundersen, and B.~E. Engelhardt, ``Bayesian non-linear
  latent variable modeling via random fourier features,'' \emph{arXiv preprint
  arXiv:2306.08352}, 2023.

\bibitem{li2024preventing}
Y.~Li, Z.~Lin, F.~Yin, and M.~M. Zhang, ``Preventing model collapse in gaussian
  process latent variable models,'' \emph{arXiv preprint arXiv:2404.01697},
  2024.

\bibitem{jung2020scalable}
Y.~Jung and J.~Park, ``Scalable hybrid hmm with gaussian process emission for
  sequential time-series data clustering,'' \emph{arXiv preprint
  arXiv:2001.01917}, 2020.

\bibitem{SASAKI20239691}
\BIBentryALTinterwordspacing
Y.~Sasaki, M.~Kawamura, and Y.~Nakamura, ``A high-speed method of segmenting
  human body motions with regular time interval sensor data based on gaussian
  process hidden semi-markov model,'' \emph{IFAC-PapersOnLine}, vol.~56, no.~2,
  pp. 9691--9696, 2023, 22nd IFAC World Congress. [Online]. Available:
  \url{https://www.sciencedirect.com/science/article/pii/S2405896323006304}
\BIBentrySTDinterwordspacing

\end{thebibliography}
\vspace{12pt}
\color{red}
\end{document}